\documentclass[conference]{IEEEtran}
\IEEEoverridecommandlockouts
\usepackage{cite}
\usepackage{subfigure}
\usepackage{amsmath,amssymb,amsfonts}
\usepackage{algorithmic}
\usepackage{graphicx}
\usepackage{textcomp}
\usepackage{xcolor}
\def\BibTeX{{\rm B\kern-.05em{\sc i\kern-.025em b}\kern-.08em
    T\kern-.1667em\lower.7ex\hbox{E}\kern-.125emX}}

\makeatletter

\let\old@ps@IEEEtitlepagestyle\ps@IEEEtitlepagestyle
\def\confheader#1{%
    \def\ps@IEEEtitlepagestyle{%
        \old@ps@IEEEtitlepagestyle%
        \def\@oddhead{\strut\hfill#1\hfill\strut}%
        \def\@evenhead{\strut\hfill#1\hfill\strut}%
    }%
    \ps@headings%
}
\makeatother

\confheader{%
\parbox{20cm} {2021 International Conference on Electronics, Communications and Information Technology (ICECIT),\\ 14–16 September 2021, Khulna, Bangladesh}
}

 \usepackage[pscoord]{eso-pic}
\newcommand{\placetextbox}[3]{
 \setbox0=\hbox{#3}
 \AddToShipoutPictureFG*{ \put(\LenToUnit{#1\paperwidth},\LenToUnit{#2\paperheight}){\vtop{{\null}\makebox[0pt][c]{#3}}}
 }
 }
 \placetextbox{.23}{0.055}{\small{978-1-6654-2363-2/21/\$31.00 ©2021 IEEE}}

\begin{document}

\title{An Automated Approach for the Recognition of Bengali License Plates\\
}

\author{\IEEEauthorblockN{Md Abdullah Al Nasim\textsuperscript{1}, Atiqul Islam Chowdhury\textsuperscript{2}, Jannatun Naeem Muna\textsuperscript{3} and Faisal Muhammad Shah\textsuperscript{4}}
\IEEEauthorblockA{\textit{Department of Computer Science and Engineering} \\
\textit{Ahsanullah University of Science and Technology\textsuperscript{1,3,4}, United International University\textsuperscript{2}}\\
Dhaka, Bangladesh \\
nasim.abdullah@ieee.org\textsuperscript{1},
achowdhury201036@mscse.uiu.ac.bd\textsuperscript{2}, muna.vns@gmail.com\textsuperscript{3} and
faisal.cse@aust.edu\textsuperscript{4}}
}

\maketitle

\begin{abstract}

Automatic Number Plate Recognition (ALPR) is a system for automatically identifying the license plates of any vehicle. This process is important for tracking, ticketing, and any billing system, among other things. With the use of information and communication technology (ICT), all systems are being automated, including the vehicle tracking system. This study proposes a hybrid method for detecting license plates using characters from them. Our captured image information was used for the recognition procedure in Bangladeshi vehicles, which is the topic of this study. Here, for license plate detection, the YOLO model was used where 81\%  was correctly predicted. And then, for license plate segmentation, Otsu's Thresholding was used and eventually, for character recognition, the CNN model was applied. This model will allow the vehicle's automated license plate detection system to avoid any misuse of it.

\end{abstract}


\section{Introduction}

Automatic Number Plate Recognition (ALPR) is a system by which the vehicles can be detected using their license plate automatically through this system without any direct human involvement. Basically, the manual process is so redundant that, by testing them randomly, the police officer must keep a record on the number of vehicles registered. ALPR represents that recognising the characters from the license plate properly and also detect them accurately \cite{i1}. ALPR system is a must one nowadays for vehicle speed measurement, automated toll collection, prevention of traffic tules violation, security controlling car parking in empty spaces, stolen vehicle search, control traffic congestion cost and many more. ALPR system basically works on object identification, preprocessing of that object image and correctly recognition of the pattern of that image. Different kinds of filters, morphological operations, transformation methods, pattern recognition are used in order to pre-process the image and then we can apply the CNN to recognize the letters correctly from the plates.

      
\begin{figure}[h]
\centering
\includegraphics[width=0.1\textwidth]{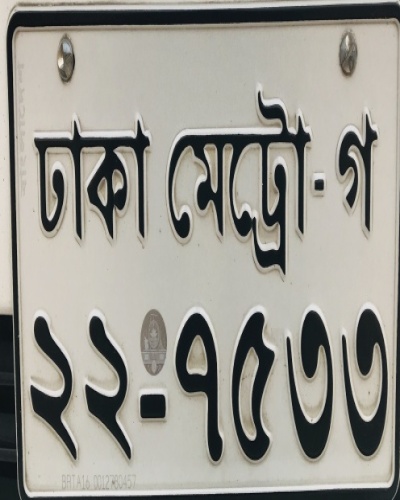}
\caption{License plate of a vehicle}
\label{f1}
\end{figure}

Bangladesh Road Transport Authority (BRTA) provides a registration number for every vehicle. The vehicle category is indicated by the vehicle class letter (KA, KHA, GHA, CHA) \cite{i2}. A lot of researches has taken place in the field of ALPR. From 2013, for every vehicle, BRTA provides a common shape of licence plate. That is called a digital license plate or Retro Reflective Licence Plate. This kind of digital plate contains two rows. At lower row, there are numbers and at the upper row, there are alphabets. In the lower row, there remain two separate parts containing six digits. The last four digits from the second part represent the vehicle number in Bengali and the first two digits from the first part represent the vehicle class. On the other hand for the upper row, a short form of the city name remains in the first word. Then for metropolitan areas, the metro is the second word and for non-metropolitan areas, there is no need to use the metro word in the license pates \cite{r1}. For the last word of the upper row that represents the category of the vehicle that means which type of vehicle is this.



\section{Related Works}

Automatic License Plate Recognition (ALPR) is a type of technology that enables computer systems to study automatically the registration number (license number) of vehicles from digital pictures. Studying automatically the registration number means
transforming the pixels of the digital image into the ASCII text of the number plate. Number plate detection is a popular work and there were so many experiments regarding this. In this section, some papers which are related and relevant to our work are being discussed.

For this research, many papers were followed which deal with Bengali Number Plate Detection and segmentation. In \cite{r4}, they applied chain code and neural network. Sobel filter, morphological operations and connected component analysis were used to extract the license plate, by scanning the image they segmented the characters and finally, recognition was done by chain code. In \cite{r1}, they used edge detection, regional localization and character segmentation for image processing and morphological operations in order to determine Bengali characters of the licence plate. They used an algorithm based on connected components.

In \cite{r2}, they worked on real-time Bengali license plate detection by using YOLOv3 algorithm foe localizing plate and recognizing the characters as well. In \cite{r3}, they used different image processing techniques to pre-process the image like resizing, image binarization, enhancement, noise filtering etc. Basically, they used a deep learning tool `The Convolutional Neural network' to recognize the plate character and they also used thresholding to extract the plate region. A paper used deep object detection convolutional neural network \cite{r6} and another one used Local binary pattern (LBP), histogram matching technique and matching bounding box technique \cite{r7}. 


In \cite{r13}, they basically showed a comparison between Gabor filter bank and fuzzified Gabor filter and the most interesting research they showed that fuzzified Gabor filter is better to detect the complex license plate than the Gabor filter bank. In \cite{r11} and \cite{r12}, they proposed a new systematic method for ALPR including a smart parking service also. In \cite{r11}, they showed that their proposed method has higher accuracy than filters Gaussian blur and filter 2D and in \cite{r12},
they preprocessed the image using sensor tools and convert the whole image to a binary image and they also used template matching, Sobel filter, dilation and median filter for ALPR.


After analysing those papers, it was found that all images need to be preprocessed and then segmentation based different techniques need to be used to achieve a good result. So, we developed a model covering all these steps in this experiment.

\section{Methodology}

For the complete detection process of characters from the license plate, three stages were split. Initially, the You Only Look Once (YOLO) paradigm was used to separate license plates from videos (mainly cars, buses on the road). Then the image was cropped using the predicted coordination of the YOLO model. And finally, using the CNN model, the digit of the segmented character was predicted. We have used our own dataset in our research for the purpose of training and testing. Our dataset comprises more than 37,700 images, which is enormous for any experiment. The dataset was divided into 70\% for training, 20\% for validation and 10\% for testing. 

The complete experiment follows some measures that provide improved performance for license plates and digits from them to be identified. The flow diagram of the proposed system of license plate recognition is shown in Fig 2.

\begin{figure}[h]
\centering
\includegraphics[width=0.4\textwidth]{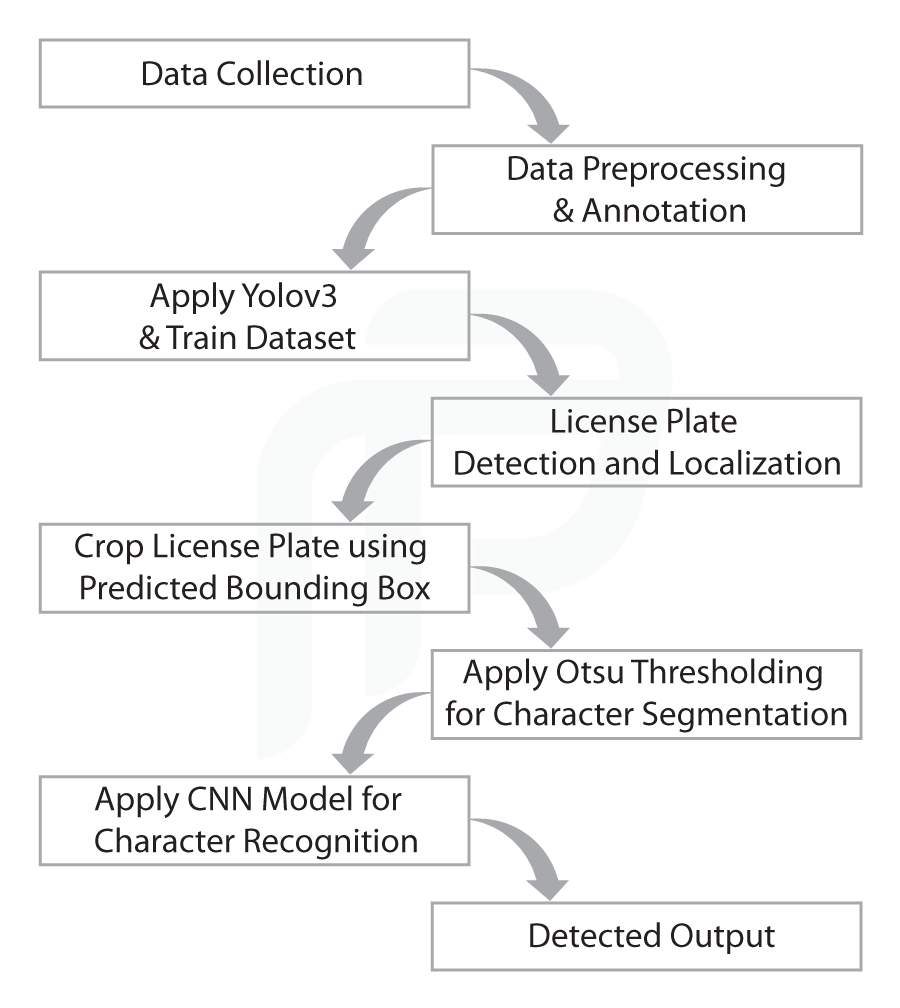}
\caption{Flow diagram of our experiment}
\label{f2}
\end{figure}

\subsection{Data Collection} 
The first step was to collect data of vehicles from roads. The images were about the front or backside of vehicles. Also, data were collected for digit recognition. The data were collected using our own mobile camera by recording video which was converted into the image frame. And from the video, different frames were separated which contain the image of the license plate. 
    
\subsection{Data Preprocessing and Annotation} 
After collecting the data, the deletion of all unclear and unimportant images was done. Then annotation was completed for the number plate of all vehicles in the first dataset. To annotate the dataset, we utilize labelimg. The dataset is manually annotated. After annotation, we use Additive Gaussian Noise, Gaussian Blur, Salt-Pepper Noise, Coarse Dropout, Contrast Normalization for license plate recognition.

\subsection{Apply YOLOv3 and Train Dataset}
Objectness was used by YOLOv3 for bounding box prediction and cost function measurement. For each bounding box using logistic regression, YOLOv3 predicts an object score. The cost function is calculated differently in YOLOv3. And this model was used for detecting the number plates which were trained on the first dataset after the annotation.

\subsection{Licence Plate Detection and Localization}
Following that, after training, the best-predicted model was used. The experiment could save the best epoch of training. After 27 epochs, the targeted detection model was found while training the data. 

\subsection{Crop Licence Plate using Predicted Bounding Box}
The saved model was used to detect the licence plate. And then, predicted bounding box coordinate was used to crop licence plates from images.

\subsection{Apply Otsu's Thresholding for Character Segmentation}
The next step was to complete the segmentation. For segmentation, Otsu’s thresholding algorithm was used. Otsu’s thresholding is an old but effective one for the segmentation process.

\subsection{Apply CNN model for Character Recognition}
A deep learning classification model, which is our CNN model, is used for character recognition. For the segmented portion of images, the model iss applied. With Keras library augmentation, Convolutional Neural Network(CNN) is used where input conv1 size is 64*64*3(rows*cols*filters), kernel size is 5*5 where size-out is 60*60*64, and conv2 size-out is 56*56*64 where we use 5*5 kernel. We use a maxpool size of 2*2 after that. The layer of Conv3 is 28*28*64 which runs up to conv7 where kernel size is 3*3, turning to flatten the 4036 layer. After flattening, the license plate is then associated with 32 dense neurons where we got the accuracy of character recognition is 99.43\%.


The YOLOv3 model was trained as the best precision after 27 epochs and saved after that epoch. In the training process, the model takes about 4.5 hours. The method is for a license plate to be detected from an image. And the updated CNN is trained at 100 epochs for character recognition. The training stage takes approximately 8.5 hours.

\section{Experimental Result}

As we mentioned in our system, for this experiment, a total of 3 phases were developed. For license plate detection, YOLOv3 was used in the first step. Then, in the second level, segmentation was completed. And the last phase was about the license plates' identification of characters. Here, all the observations of these three phases are illustrated.

Licence plate detection using YOLOv3 had an output accuracy of 0.81 mAP (Mean Average Precision). That means 81\% guess/prediction is correct. After training with a lot of data, the test phase showed us a better result for license plate detection. Images from different angles were used for testing purpose. And our proposed model worked much better regarding those issues. Figure 3 and 4 show the detection of a license plate from an image.

\begin{figure}[h]
\centering
\includegraphics[width=0.25\textwidth]{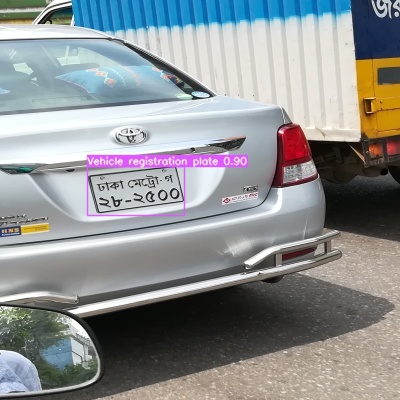}
\caption{License plate recognition output (120 degree angle)}
\label{f3}
\end{figure}

\begin{figure}[h]
\centering
\includegraphics[width=0.25\textwidth]{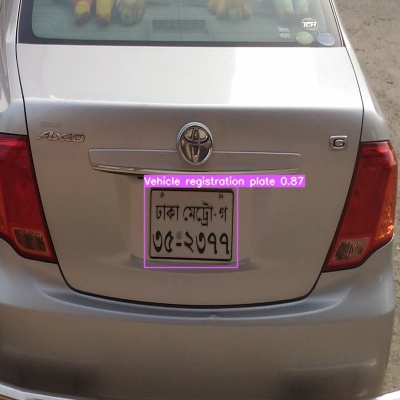}
\caption{License plate recognition output (90 degree angle)}
\label{f4}
\end{figure}

\begin{figure*}[!ht]
\centering
\includegraphics[width=0.7\textwidth]{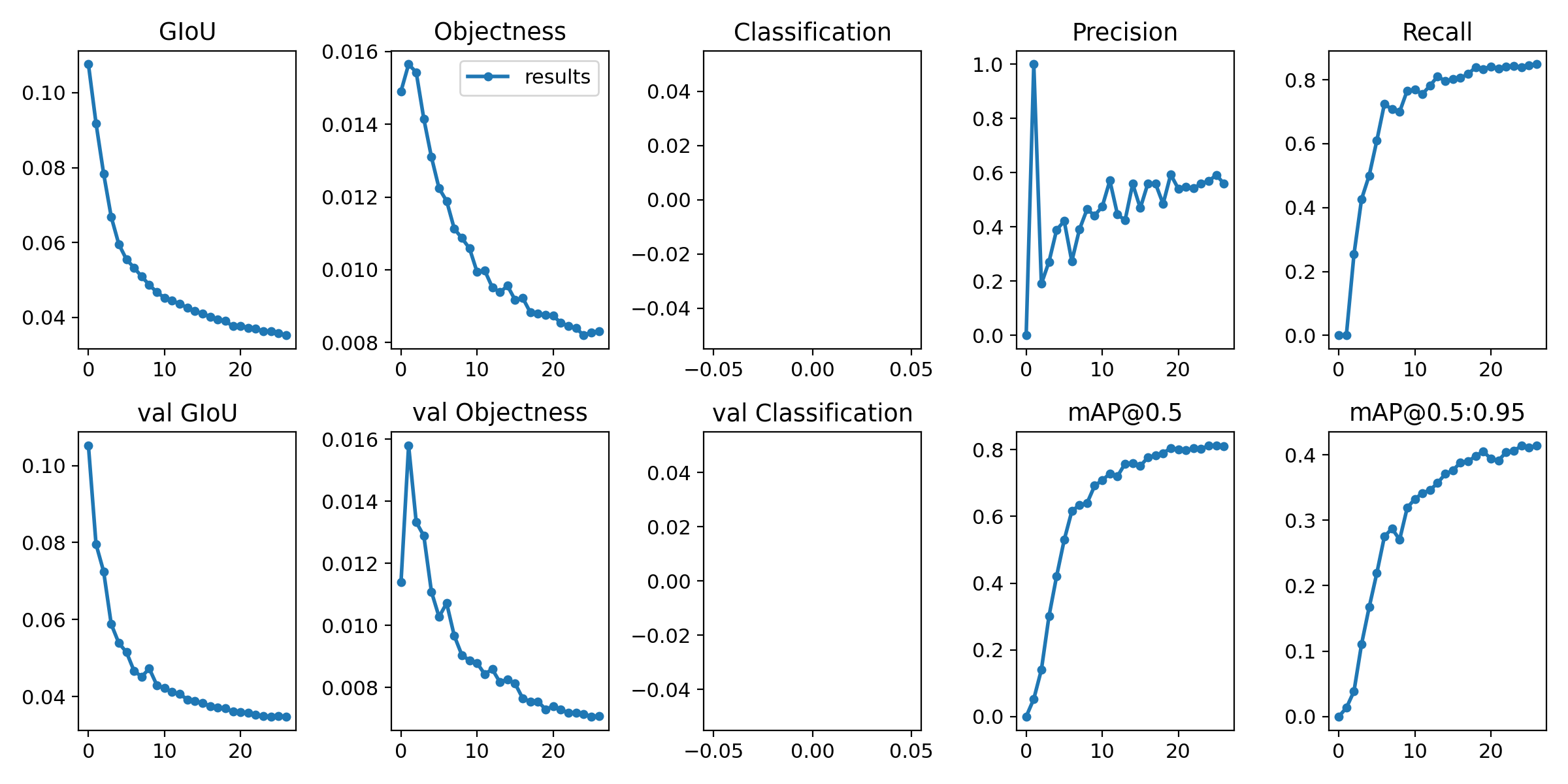}
\caption{Analysis of YOLOv3 Model for License Plate Detection}
\label{f7}
\end{figure*}

These figures show that the YOLOv3 model provides a good output for the detection of number plate at any angles. Besides this, from figure 5, there are many different terms which shows a graphical view of the detection process. GIoU measures the overlap between two boundaries of Bengali license plate. Here, loss degrades in every step which helps us to lead a better accuracy. In figure 5 of GIoU, we have predicted the bounding box with label bounding box figure with shape and orientation of coverage area. It determines the localization of Bengali license plate which is differentiable.  In this graph, objectness means how accurately we bounded the box which contains license plate. Here our estimated object detection of binary cross-entropy loss is decreasing which means our predicted license plate detection's bounding box works better. The classification works to classify how many objects we detect here. We just detect Bengali license plates in this study, thus no classification is required. In 3770 test images, there are 3580 images where we find there are license plate and predict that there are 3270 images were detected correctly. Here we accurately predict 3270 images and 170 images which can not detect license plate accurately. Prediction of bounding license plate detection and actual bounding box for license plate detection which is called Mean Average Precision (mAP) is 0.81.

After that, the images of license plates were cropped and segmentation was applied for the better look of every character on the plate. The last phase is character recognition from those plates. Character recognition is not an easy step as sometimes images are blurry, sometimes cropped etc. So these issues need to be maintained for any experiment. Then we utilized our modified CNN to recognize the characters, which had a validation accuracy of 99.43\%. We have experimented different ratio which are shown in Table I.


\begin{figure}[ht]
    \centering
    \subfigure[]
    {
        \includegraphics[width=1.2in]{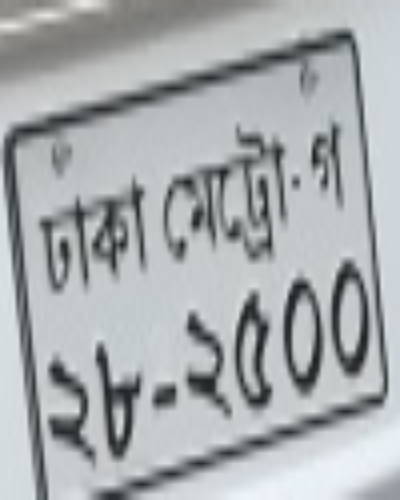}
        \label{fig:first_sub}
    }
    \subfigure[]
    {
        \includegraphics[width=1.2in]{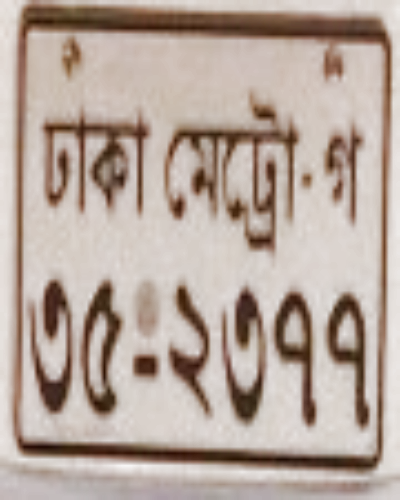}
        \label{fig:second_sub}
    }
    \caption{Segmentation of the license plate image}
    \label{f5}
\end{figure}


\begin{figure}[!ht]
    \centering
    \subfigure[]
    {
        \includegraphics[width=1.2in]{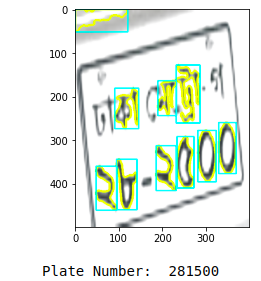}
        \label{fig:first_sub}
    }
    \subfigure[]
    {
        \includegraphics[width=1.2in]{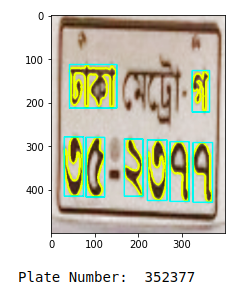}
        \label{fig:second_sub}
    }
    \caption{Recognition of the characters from the license plates}
    \label{f6}
\end{figure}

\begin{table}[!ht]
\caption{Ratio of Training \& Testing Data}
\begin{center}
\begin{tabular}{|l|r|}\hline
\textbf{Ratio of Training \& Testing Data} & \textbf{Accuracy (\%)}\\ \hline
70:30 & 91.72 \\
80:20 & 99.43  \\
85:15 & 97.14 \\
\hline
\end{tabular}
\label{table:1}
\end{center}
\end{table}

Finally, it is obvious that the identification of license plates and characters fits best with our model. It was a tough problem for the detection system, as we used our own dataset from the Bangladesh vehicle. But despite those problems, our hybrid model shows strong performance.

\section{Conclusion}

Automatic Number Plate detection is a whole package of capturing the license plate from the vehicle and to recognize it accurately. Basically, the purpose of this system is to detect the Bengali licence plate to identify the vehicles properly who violates the rules of the road. The whole procedure was divided into three subprocedures. First YOLOv3 model was used for the detection of a license plate from the selected images. Then predicted bounding box was used for cropping the images of the license plate. For segmenting the character, Otsu's thresholding method was applied. Then our own modified deep learning classification CNN model was used for character recognition. At the end, we can say that this research provides a better result for the overall recognition of license plate and character. As a large number of different vehicles are now added and to handle them properly, providing this device is vital for every corner of the city, the future work of this scheme will be much brighter.

\bibliographystyle{IEEEtran}
\bibliography{bibliography}

\end{document}